\documentclass[conference]{IEEEtran}
\IEEEoverridecommandlockouts
\usepackage{cite}
\usepackage{amsmath,amssymb,amsfonts}
\usepackage{hyperref}

\usepackage{mathtools} 
\usepackage{booktabs} 
\usepackage{tikz} 
\usepackage{booktabs}
\usepackage{multirow}
\usepackage{amsthm}

\usepackage{algpseudocode}
\usepackage{algorithmicx}
\usepackage{adjustbox}

\usepackage{xcolor}

\usepackage{amsmath}

\usepackage{algorithm}
\usepackage{algorithmicx}
\usepackage{subcaption}
\usepackage{amsfonts}

\begin{document}

\title{A Fast Interpretable Fuzzy Tree Learner }

\author{\IEEEauthorblockN{Javier Fumanal-Idocin}
\IEEEauthorblockA{\textit{School of Computer Science} \\
\textit{and Electronic Engineering} \\
\textit{University of Essex}\\
Colchester, United Kingdom \\
j.fumanal-idocin@essex.ac.uk}
\and
\IEEEauthorblockN{Raquel Fernandez-Peralta}
\IEEEauthorblockA{\textit{Department of Mathematics} \\
\textit{Slovak Academy of Sciences}\\
Bratislava, Slovakia \\
raquel.fernandez@mat.savba.sk}
\and
\IEEEauthorblockN{Javier Andreu-Perez}
\IEEEauthorblockA{\textit{School of Computer Science} \\
\textit{and Electronic Engineering} \\
\textit{University of Essex}\\
Colchester, United Kingdom \\
j.andreu-perez@essex.ac.uk}
}
\maketitle

\begin{abstract}
Fuzzy rule-based systems have been mostly used in interpretable decision-making because of their interpretable linguistic rules. However, interpretability requires both sensible linguistic partitions and small rule-base sizes, which are not guaranteed by many existing fuzzy rule-mining algorithms. Evolutionary approaches can produce high-quality models but suffer from prohibitive computational costs, while neural-based methods like ANFIS have problems retaining linguistic interpretations. In this work, we propose an adaptation of classical tree-based splitting algorithms from crisp rules to fuzzy trees, combining the computational efficiency of greedy algoritms with the interpretability advantages of fuzzy logic. This approach achieves interpretable linguistic partitions and substantially improves running time compared to evolutionary-based approaches while maintaining competitive predictive performance. Our experiments on tabular classification benchmarks proof that our method achieves comparable accuracy to state-of-the-art fuzzy classifiers with significantly lower computational cost and produces more interpretable rule bases with constrained complexity.
Code is available in: \url{https://github.com/Fuminides/fuzzy_greedy_tree_public}
\end{abstract}

\begin{IEEEkeywords}
Explainable Artificial Intelligence, Fuzzy Logic, Classification, Rule Mining \end{IEEEkeywords}

\section{Introduction}
Fuzzy rule-based systems have been popular for explainable artificial intelligence (XAI) as they rely on logic-based reasoning that is easily understood by humans \cite{mendel2024explainable,rudin2022interpretable}. However, this explainability is limited by the complexity of the system. For example, a system composed of hundreds of rules is much harder to understand than one with only a handful \cite{gacto2011interpretability}. Therefore, besides the use of highly performant classifiers such as FURIA \cite{huhn2009furia}, there has also been particular research interest in a subset of fuzzy rule-based classifiers (FRBCs) that constrain the resulting rule base according to the end user's desired complexity \cite{fumanal2023artxai, gonzalez2024rule}.

Two main approaches have been proposed to address this challenge. Architecturally constrained methods, such as Adaptive Network-based Fuzzy Inference System (ANFIS) \cite{jang1993anfis} and Fuzzy Rule Reasoner (FRR) \cite{fumanal2025compact}, use neural architectures where the number of rules and conditions are pre-specified. Evolutionary optimization approaches \cite{alcala2011fuzzy,oh2004hybrid, fumanal2024ex} can incorporate complexity constraints into the optimization process, just as gradient-based methods enforce sparsity through loss functions \cite{mctavish2022fast}. However, this approach has limitations: constraints are often implemented as soft preferences that may still result in complex models, and even when successful, the computational cost typically exceeds that of standard classifiers, discouraging practitioners from using FRBCs. 

The most popular method for performing crisp rule-based classification has been to use tree structures with Gini-conditioned splits. This has provided a computationally efficient and accurate way to perform rule learning, establishing CART \cite{breiman2017classification} as a widely adopted classification algorithm. However, CART has significant limitations from an interpretability perspective. First, the number of leaf nodes can be quite high, making the overall decision structure difficult to understand. Second, CART uses hard splits that create arbitrary boundaries (e.g., ``age $<$ 45.7''), which do not reflect the gradual transitions often present in human reasoning. Finally, CART provides no mechanism for incorporating existing expert knowledge into the learning process, potentially producing rules that conflict with established domain knowledge or are difficult for practitioners to trust. This leaves room for opportunity for fuzzy systems, where conditions can be defined to take into account the intuitions of practitioners through meaningful linguistic partitions \cite{zadeh1975concept}, and can potentially recover complex decision surfaces using fewer rules \cite{fernandez2025crisp}. However, the task of finding optimal fuzzy partitions while keeping interpretability is not trivial. Fuzzy systems that rely on predefined fuzzy partitions are computationally efficient, but this strategy may produce suboptimal decision boundaries when the class distribution does not align well with this pre-existing semantics.

Because of this, in this paper we propose a fuzzy tree learner algorithm with a greedy approach, designed for computational efficiency. The algorithm is designed to be competitive against straightforward genetic approaches while maintaining good interpretability by imposing strict constraints on the number of rules and conditions. We also propose a lightweight optimization framework that tunes fuzzy partition parameters to maximize class separability while maintaining computational tractability and interpretability guarantees. Unlike evolutionary approaches that require thousands of fitness evaluations, our method employs efficient search strategies to keep computational complexity low.

The rest of the paper goes as follows: in Section \ref{sec:related} we discuss relevant literature regarding explainability in rule-based systems. Then, we present our learning approach in Section \ref{sec:methodology}. Subsequently, in Section \ref{sec:experiments} we discuss our experimentation and show our empirical results. Finally, in Section \ref{sec:conclusions}, we give our conclusions and future lines for this work.

\section{Background and Related Work} \label{sec:related}

\subsection{Complexity Accuracy Trade-off in rule-based learning}
Rule-based learning methods need to balance model complexity and predictive accuracy \cite{rudin2022interpretable}. As the number of rules and conditions increases, the model can capture more specific patterns in the data, typically leading to improved classification performance. However, this comes at the cost of interpretability: a rulebase with hundreds of rules becomes essentially a black box, defeating the primary purpose of using rule-based systems for explainable AI \cite{rudin2019stop,arrieta2020explainable}. This trade-off is particularly critical in domains such as healthcare \cite{chua2023tackling}, finance \cite{erion2022cost}, and legal decision-making, where stakeholders must not only trust the model's predictions but also understand the reasoning behind them \cite{lipton2018mythos}. Consequently, finding the optimal balance between accuracy and complexity remains a central challenge in rule-based machine learning \cite{molnar2020interpretable}.

\subsection{Different crisp and fuzzy rule-learning strategies}

Classical tree-based approaches, like CART \cite{breiman2017classification} and C4.5 \cite{quinlan2014c4}, remain among the most popular crisp rule-based algorithms due to their greedy, top-down induction strategy. These algorithms achieve $O(n \log n)$ training complexity by making locally optimal splits at each node without global backtracking. The resulting decision trees naturally provide complexity control through depth limits and pruning mechanisms, and the hierarchical structure offers intuitive decision pathways. More novel approaches employ gradient-based learning to improve scalability \cite{wang2021scalable, wang2023learning, qiao2021learning}, and Bayesian approaches that try to minimize the size of the rulebase with guarantees \cite{wang2017bayesian}.


For fuzzy systems, gradient-based approaches has been very popular. The ANFIS \cite{jang1993anfis} is perhaps the most well-known example, combining neural network learning capabilities with fuzzy reasoning. ANFIS employs a hybrid learning algorithm that uses backpropagation to tune membership function parameters in the premise part and least-squares estimation for the consequent parameters. However, this fine-tuning comes at the cost of interpretability: the learned membership functions often lose their linguistic meaning, with overlapping or irregular shapes. Additionally, ANFIS requires the number of rules to be specified \textit{a priori}, and for problems with multiple input variables, the curse of dimensionality quickly leads to an exponential explosion in the number of rules needed. More recent advances in gradient-based fuzzy learning use classic Mamdani fuzzy inference \cite{fumanal2025compact} with a neural architecture that support rule-learning without curse of dimensionality and retains interpretability of linguistic partitions. Evolutionary approaches offer greater flexibility in exploring the space and can simultaneously optimize rule structure, membership function parameters, and the number of rules \cite{fumanal2024ex, alcala2011fuzzy}. However, the computational cost remains prohibitive for many applications, as evolutionary algorithms typically require thousands of fitness evaluations.

The approach presented in this paper aims to bridge these paradigms by adapting the computational efficiency of tree-based algorithms to produce genuinely fuzzy rule bases with linguistic interpretability, while maintaining strict complexity control comparable to classical decision trees.

\section{Greedy Fuzzy Rule Tree Induction}  \label{sec:methodology}


We propose a novel fuzzy tree learning algorithm that efficiently constructs interpretable rule bases through greedy condition selection from pre-defined linguistic partitions. Unlike traditional decision tree algorithms that search for optimal split points in continuous feature spaces, our approach leverages existing fuzzy membership functions to build rules with natural linguistic interpretability (e.g., ``IF `Age' is \textit{High} AND `Income' is \textit{Medium}...''). So, as long as the linguistic partitions are interpretable by the user, the resulting rules should be interpretable as well.

The key main characteristics of our approach compared to other fuzzy rule learners are:

\begin{enumerate}
    \item Greedy Condition Search: At each node, the algorithm performs a heuristic search over all available fuzzy conditions, selecting the single condition that maximizes impurity reduction. This greedy strategy achieves computational efficiency comparable to classical tree learners, like CART.
    
    \item Conjunctive Rule Building: Rules are constructed incrementally by adding conditions to existing nodes. Each path from root to leaf represents a complete fuzzy rule of the form: IF ($X_1$ is $A_1$) AND ($X_2$ is $A_2$) AND ... THEN Class. Internal nodes can also be considered if leafs where not significantly fired.
    
    \item Multi-way Branching: Unlike binary tree algorithms that partition data into two subsets at each split, our approach uses natural multi-way splits 
    using the given fuzzy partitions. A single node can spawn multiple children, each corresponding to a different linguistic value of the selected feature. We do not use negation conditions, as they significantly hinder rule interpretability.
\end{enumerate}

\subsection{Fuzzy Split Quality Measure}

To evaluate the quality of fuzzy conditions, we extend classical impurity metrics to handle fuzzy memberships. For a given rule $R$ and dataset $D$, each instance $x_i \in D$ has a membership degree $\mu_R(x_i) \in [0,1]$ indicating the extent to which it satisfies $R$. 

The fuzzy Gini impurity for a node with rule $R$ is computed as \cite{belhadj2021fuzzy}:
\begin{equation}
    Gini_{fuz}(R) = 1 - \sum_{c \in Classes} \left(\frac{\sum_{i: y_i=c} \mu_R(x_i)}{\sum_{i} \mu_R(x_i)}\right)^2
\end{equation}
where $y_i$ is the class label of instance $x_i$. This formulation weights each instance's contribution to class proportions by its membership degree to the rule. The impurity gain when adding a new condition $(X_j \text{ is } A_j)$ to rule $R$ is:
\begin{equation}
    Gain(R, X_j, A_j) = Gini_{fuz}(R) - Gini_{fuz}(R \land (X_j \text{ is } A_j))
\end{equation}

\subsection{Algorithm Description}

Algorithm \ref{alg:gfrt} details the core tree induction procedure. The algorithm operates recursively, building the tree in a depth-first manner. At each node:

\begin{enumerate}
    \item Termination Check (lines 4-6): If stopping criteria are met (e.g., minimum membership threshold, maximum depth, class purity), the node becomes a leaf and its predictions correspond to its confidence for each class.
    
    \item Evaluate Existing Children (lines 9-15): For nodes with existing children, the algorithm considers if further searching in those children would improve the partition.
    
    \item Search New Conditions (lines 16-27): The algorithm evaluates all valid fuzzy conditions from unused features. For each feature $F$ and each linguistic term $A$ in its partition, it computes the impurity gain of adding $(F \text{ is } A)$ to the current rule. The condition $(F,A)$ is valid if feature $F$ has not been used in the path from root to current node, preventing contradictory conditions like $(Age \text{ is } High) \land (Age \text{ is } Low)$.
    
    \item New Node Selection (lines 21-24): Among all candidate conditions (both from existing children and new features), the algorithm selects the single best condition that maximizes impurity gain.
    
    \item Expansion (lines 28-30): If the best gain exceeds threshold $\theta$, a new child node is created with the extended rule, and the procedure recurses. The threshold $\theta$ serves as a pruning mechanism, preventing splits that provide minimal improvement.
\end{enumerate}

\begin{algorithm}[H]
\caption{Greedy Fuzzy Rule Tree Induction}
\label{alg:gfrt}
\footnotesize
\begin{algorithmic}[1]
\Procedure{GrowTree}{$Node, Data, FuzzyPartitions$}
\State $Rule \gets Node.Rule$ \Comment{Current conjunctive fuzzy rule}
\State $Memberships \gets$ ComputeMemberships($Data$, $Rule$)
\If{StoppingCriterion($Node, Data, Memberships$)}
    \State $Node.Class \gets$ MajorityClass($Data, Memberships$)
    \State \Return
\EndIf
\State $BestCondition \gets \emptyset$
\State $BestGain \gets -\infty$
\For{each child $C$ in $Node.Children$}
    \State $gain, condition \gets$ EvaluateBestSplit($Rule$, $C$, $Data$)
    \If{$gain > BestGain$}
        \State $BestGain \gets gain$
        \State $BestCondition \gets condition$
    \EndIf
\EndFor
\For{each feature $F$ in $Features$}
    \If{$F$ not used in path to $Node$} \Comment{Avoid contradictions}
        \For{each linguistic term $A$ in $FuzzyPartitions[F]$}
            \State $NewRule \gets Rule \land (F \text{ is } A)$
            \State $Gain \gets$ ImpurityGain($Data$, $Rule$, $NewRule$)
            \If{$Gain > BestGain$}
                \State $BestGain \gets Gain$
                \State $BestCondition \gets (F, A)$
            \EndIf
        \EndFor
    \EndIf
\EndFor
\If{$BestCondition \neq \emptyset$ \textbf{and} $BestGain > \theta$}
    \State $ChildNode \gets$ CreateChildNode($BestCondition$)
    \State \Call{GrowTree}{$ChildNode, Data, FuzzyPartitions$}
\EndIf
\EndProcedure
\end{algorithmic}
\end{algorithm}

\subsection{Computational Complexity Analysis}

The proposed greedy fuzzy tree learning algorithm achieves a time complexity of $\mathcal{O}(n \cdot m \cdot k \cdot d)$, where $n$ is the number of training instances, $m$ is the number of features, $k$ is the average number of linguistic terms per feature, and $d$ is the maximum tree depth. At each node in the tree, the algorithm evaluates all possible fuzzy conditions across $m$ features and their corresponding $k$ linguistic partitions, computing membership degrees and weighted Gini impurity for $n$ instances, resulting in $\mathcal{O}(n \cdot m \cdot k)$ operations per node. The greedy construction strategy makes one pass through the candidate splits without backtracking, limiting the total number of node evaluations to the tree depth $d$. This represents a substantial improvement over evolutionary approaches that typically require $\mathcal{O}(p \cdot g \cdot n \cdot m \cdot r)$ operations for $p$ population size, $g$ generations, and $r$ rules, often resulting in thousands of complete model evaluations.

\section{Fuzzy Partition Optimization}

\subsection{Partition Representation}

Each linguistic term is represented by a trapezoidal membership function $\mu(x; a, b, c, d)$, defined by four parameters where $a \leq b \leq c \leq d$:

\begin{equation}
\mu(x; a,b,c,d) = \begin{cases} 
0 & \text{if } x < a \\
\frac{x-a}{b-a} & \text{if } a \leq x < b \\
1 & \text{if } b \leq x \leq c \\
\frac{d-x}{d-c} & \text{if } c < x \leq d \\
0 & \text{if } x > d
\end{cases}
\end{equation}

The key challenge in optimizing these parameters is ensuring that validity constraints are maintained throughout optimization: each trapezoid must satisfy $a \leq b \leq c \leq d$, and the trapezoids must maintain interpretable ordering (\textit{Low} $<$ \textit{Medium} $<$ \textit{High}). 

To enable unconstrained optimization while guaranteeing validity by construction, we developed a novel interleaved encoding scheme. Rather than directly optimizing the trapezoid parameters, we encode them as incremental differences in a carefully chosen order. The encoding interleaves parameters from adjacent trapezoids, ensuring that touching points between consecutive linguistic terms are encoded together. For example, three partitions with parameters $\text{Low}[a_1, b_1, c_1, d_1]$, $\text{Medium}[a_2, b_2, c_2, d_2]$, and $\text{High}[a_3, b_3, c_3, d_3]$, are represented in the interleaved order as:

\begin{equation}
\begin{split}
\mathbf{encoded} = [a_1,\, b_1-a_1,\, c_1-b_1,\, a_2-c_1,\, d_1-a_2,\, \\
b_2-d_1,\, c_2-b_2,\, a_3-c_2,\, d_2-a_3,\, b_3-d_2,\, c_3-b_3,\, d_3-c_3]
\end{split}
\end{equation}

By representing each position (except the first) as the positive difference from the previous accumulated value, we ensure:
\begin{enumerate}
    \item All accumulated values are monotonically increasing by construction
    \item Each trapezoid automatically satisfies $a \leq b \leq c \leq d$
    \item Trapezoids maintain interpretable ordering
    \item Full domain coverage is guaranteed
    \item No constraint violations are possible during optimization
\end{enumerate}

The decoding process accumulates these increments, extracts parameters in the interleaved order, and normalizes to the feature domain.

\subsubsection{Optimization Objective}

We optimize partitions to maximize the \textit{separability index}, which measures how well each fuzzy partition concentrates samples of the same class:

\begin{equation}
SI = \sum_{v=1}^{V} \sum_{c=1}^{C} \frac{\left[\sum_{i=1}^{N} \mu_v(x_i) \cdot \mathbb{I}(y_i = c)\right]^2}{\sum_{i=1}^{N} \mu_v(x_i)}
\end{equation}
where $v$ indexes linguistic terms, $c$ indexes class labels, $\mu_v(x_i)$ is the membership degree of sample $i$ to term $v$, and $\mathbb{I}(y_i = c)$ is the indicator function for class membership. Higher values indicate better class separation in the fuzzy space, as partitions that concentrate same-class samples together yield larger numerators while diverse class mixing yields smaller values.

\subsection{Search Strategy}

We optimize one parameter at a time cyclically, using progressively smaller step sizes: $\{0.10, 0.05, 0.02\} \times (x_{\max} - x_{\min})$. At each iteration, the algorithm tests both directions ($\pm$) for each of the three critical parameters and accepts improvements. The algorithm converges when no improvement is found with the smallest step size.

This typically requires only 30--84 evaluations per feature and while it may converge to local optima, empirical results show it often finds better solutions than grid search due to its adaptive exploration of the parameter space.

\subsection{Integration with Tree Learning}

The optimized partitions are generated once during the training phase using only the training data within each cross-validation fold. The tree learning procedure then uses these optimized partitions in place of the default quantile-based partitions, with all other algorithmic components remaining unchanged.

\section{Experimental Setup}
\label{sec:experiments}

\subsection{Datasets and evaluation}

We selected 10 benchmark classification datasets from the UCI Machine Learning Repository~\cite{kelly2025uci}, chosen to represent diverse application domains and exhibit varying characteristics in terms of size and dimensionality. Table~\ref{tab:datasets} summarizes the properties of each dataset. Results are reported using 5-fold cross validation and all features are normalized by subtracting their mean and dividing for the standard deviation. Results are reported using the classification accuracy metric. To measure complexity in the resulting rulebases we use the average rule size, the number of rules per classifier, and the total number of conditions in the rulebase.

\begin{table}[t]
\centering
\caption{Characteristics of datasets used in our experiments.}
\label{tab:datasets}
\begin{tabular}{lrrrr}
\toprule
\textbf{Dataset} & \textbf{Instances} & \textbf{Features} & \textbf{Classes} & \textbf{Domain} \\
\midrule
Appendicitis     & 106   & 7  & 2  & Medical \\
Australian       & 690   & 14 & 2  & Financial \\
Dermatology      & 366   & 34 & 6  & Medical \\
Hepatitis        & 155   & 19 & 2  & Medical \\
Pima             & 768   & 8  & 2  & Medical \\
Ring             & 7,400 & 20 & 2  & Synthetic \\
Saheart          & 462   & 9  & 2  & Medical \\
Spambase         & 4,601 & 57 & 2  & Text/Email \\
Wine             & 178   & 13 & 3  & Chemical \\
Zoo              & 101   & 16 & 7  & Biology \\
\bottomrule
\end{tabular}
\end{table}

\subsection{Baseline Methods}

We compared approach against five established rule-learning methods representing different algorithmic paradigms:

\subsubsection{Crisp Rule-Based Classifiers}

\begin{itemize}
    \item Classification and Regression Trees~\cite{breiman2017classification}: the foundational decision tree algorithm that uses Gini impurity for split selection. We used the scikit-learn~\cite{scikit-learn} implementation with cost complexity pruning parameters tried at $\{0.0, 0.001, 0.003\}$.
    
    \item C4.5 \cite{quinlan2014c4}: An extension of ID3 that employs information gain ratio for split selection and includes pessimistic pruning. Implemented via scikit-learn with and maximum depths of \{5, 10\}.
    
    \item RIPPER \cite{cohen1995fast}: Repeated Incremental Pruning to Produce Error Reduction, a separate-and-conquer rule learning algorithm that constructs ordered rule lists. We used the \texttt{wittgenstein} Python library implementation with default parameters.
\end{itemize}

\subsubsection{Fuzzy Rule-Based Classifiers}

\begin{itemize}
    \item FRR~\cite{fumanal2025compact}: Fuzzy Rule Reasoner, a recent gradient-based approach that learns Mamdani fuzzy rule bases while preserving linguistic interpretability through a neural architecture that avoids the curse of dimensionality and non-differentiability issues.
    
    \item Genetic Fuzzy Classifier (FGA)~\cite{fumanal2024ex}: An evolutionary algorithm-based fuzzy classifier, implemented  using the Ex-Fuzzy library~\cite{fumanal2024ex}.
\end{itemize}

\subsection{Fuzzy Partition Configuration}

To ensure fair comparison across all fuzzy methods, we employed consistent fuzzy partitions throughout the experiments, following the quantile-based approach described in the FRR methodology~\cite{fumanal2025compact}. This ensures that differences in performance come from the rule-learning strategy rather than fuzzy partition choices. For that, each continuous feature was partitioned using three trapezoidal membership functions with linguistic labels \textit{Low}, \textit{Medium}, and \textit{High}. 

The membership functions are defined using quantile-based parameters:
    \begin{itemize}
        \item Low: Trapezoid $(Q_0, Q_0, Q_1, Q_2)$
        \item Medium: Trapezoid $(Q_1, (Q_1+Q_2)/2, (Q_2+Q_3)/2, Q_3)$
        \item High: Trapezoid $(Q_2, Q_3, Q_4, Q_4)$
    \end{itemize}
    where $Q_i$ represents the $i$-th quantile at percentages $\{0, 20, 40, 60, 80, 100\}$ of the training data.

\section{Experimental Results}
\label{sec:results}

\subsection{Classification Performance}

Table~\ref{tab:accuracy} presents the classification accuracy results comparing our proposed FGRT with established baseline methods across 10 benchmark datasets using 5-fold cross-validation. FGRT with default quantile partitions achieves 76.92\% mean accuracy, which is competitive with FGA (76.91\%) and matches RIPPER (76.93\%), both established rule-learning algorithms. While FGRT does not reach the accuracy levels of CART (81.77\%) or C4.5 (83.57\%), this 4.85 percentage point gap is expected given that FGRT operates under stricter interpretability constraints—using predefined linguistic partitions rather than optimizing split points, and maintaining interpretable fuzzy rules rather than creating potentially hundreds of crisp leaf nodes.

\begin{table}[t]
\centering
\caption{Classification accuracy (\%) across all datasets using 5-fold cross-validation. FGRT denotes our Fuzzy Greedy Rule Tree with default quantile partitions, FGRT$^*$ uses optimized separability-based partitions.}
\label{tab:accuracy}
\adjustbox{width=\linewidth}{
\begin{tabular}{l|cc|ccccc}
\toprule
\textbf{Dataset} & \textbf{FGRT} & \textbf{FGRT$^*$} & \textbf{FRR} & \textbf{CART} & \textbf{C4.5} & \textbf{RIPPER} & \textbf{FGA} \\
\midrule
Appendicitis     & 80.22 & 81.17 & 86.36 & 72.55 & 74.50 & 84.03 & 90.91 \\
Australian       & 86.09 & 85.65 & 83.91 & 81.16 & 83.04 & 83.04 & 85.51 \\
Dermatology      & 68.45 & 66.76 & 80.05 & 86.87 & 86.87 & 84.13 & 50.00 \\
Hepatitis        & 83.75 & 85.00 & 90.00 & 83.75 & 90.00 & 66.15 & 81.25 \\
Pima             & 65.10 & 65.10 & 72.59 & 69.15 & 69.41 & 70.08 & 74.68 \\
Ring             & 59.86 & 83.70 & 84.41 & 88.15 & 88.82 & 60.55 & 68.58 \\
Saheart          & 65.37 & 65.37 & 70.96 & 60.37 & 84.74 & 64.05 & 66.67 \\
Spambase         & 88.71 & 90.12 & 78.78 & 90.80 & 92.23 & 87.33 & 88.48 \\
Wine             & 94.41 & 96.62 & 97.22 & 89.87 & 92.08 & 86.20 & 91.67 \\
Zoo              & 77.24 & 77.24 & 66.85 & 95.05 & 94.05 & 83.75 & 71.43 \\
\midrule
\textbf{Mean}    & 76.92 & 79.67 & 81.11 & 81.77 & 83.57 & 76.93 & 76.91 \\
\bottomrule
\end{tabular}}
\end{table}

Notably, FGRT demonstrates strong dataset-specific performance that highlights the effectiveness of fuzzy partitioning. On \textit{australian}, FGRT achieves 86.09\%---the highest accuracy among all methods tested. On \textit{wine}, FGRT reaches 94.41\%, surpassing CART (89.87\%), C4.5 (92.08\%), RIPPER (86.20\%), and FGA (91.67\%), demonstrating that when fuzzy partitions effectively capture decision boundaries, FGRT can outperform traditional crisp rule learners. Similarly, on \textit{spambase}, FGRT achieves 88.71\%, remaining competitive with more complex approaches while maintaining interpretability.

Compared to FRR, which achieves 81.11\% mean accuracy through gradient-based optimization, FGRT's 4.19 percentage point lower accuracy represents a reasonable trade-off considering the algorithmic differences: FRR requires multiple training epochs, careful hyperparameter tuning, GPU acceleration, and backpropagation through fuzzy inference layers, while FGRT uses a single-pass greedy construction that completes in seconds on standard CPU hardware.

\subsection{Impact of Partition Optimization}

\begin{figure}[t]
    \centering
    \includegraphics[width=\linewidth]{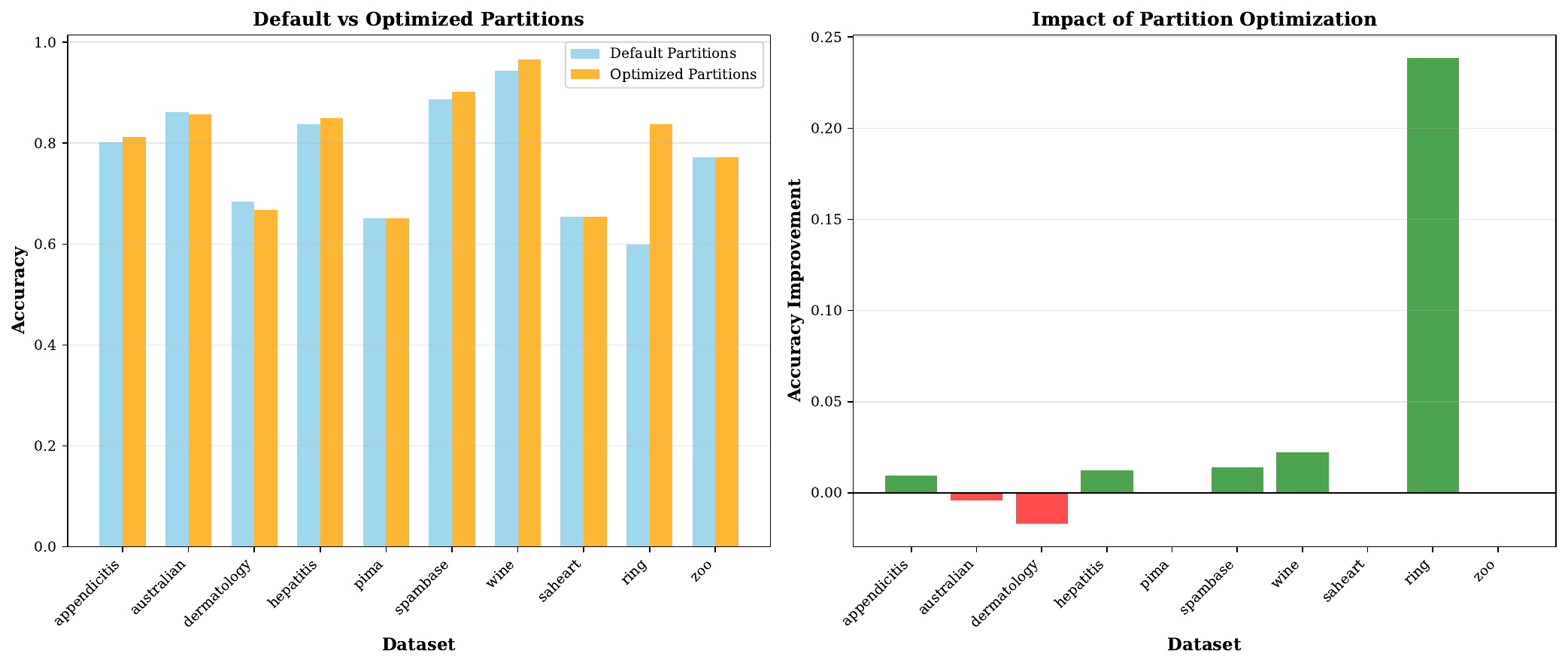}
    \caption{Performance improvements using pre-configured or optimized fuzzy sets. Left: Accuracy comparison showing default (blue) versus optimized (orange) partitions. Right: absolute accuracy improvements.}
    \label{fig:opt_impact}
\end{figure}

We use our propose fuzzy partition optimization to investigate whether fuzzy partition quality limits FGRT's performance. We denote this variant as FGRT$^*$. Figure~\ref{fig:opt_impact} visualizes the impact of this optimization across all datasets, comparing default quantile-based partitions with optimized separability-based partitions.

As shown in Table~\ref{tab:accuracy}, FGRT$^*$ with optimized partitions achieves 79.67\% mean accuracy, representing a +2.75 percentage point improvement over the default configuration. This optimization improved accuracy on 5 out of 10 datasets, with three datasets (\textit{pima}, \textit{saheart}, \textit{zoo}) showing no change, indicating that default quantile partitions were already well-aligned with their class structures. Two datasets show slight decreases: \textit{dermatology} ($-1.7$ pp) and \textit{australian} ($-0.4$ pp). This shows that for datasets where default quantile partitions already provide near-optimal decision boundaries, further optimization can introduce minor overfitting to the training fold's specific class distribution.

\subsection{Model Complexity}

Beyond predictive accuracy, interpretability requires models with manageable structural complexity. Table~\ref{tab:complexity} presents complexity metrics averaged across all 10 datasets, quantifying the size and structure of the learned rule bases.

\begin{table}[t]
\centering
\caption{Average model complexity metrics across all 10 datasets. Lower values indicate simpler, more interpretable models.}
\label{tab:complexity}
\adjustbox{width=\linewidth}{
\begin{tabular}{lcccccc}
\toprule
\textbf{Metric} & \textbf{FGRT} & \textbf{FRR} & \textbf{CART} & \textbf{C4.5} & \textbf{RIPPER} & \textbf{FGA} \\
\midrule
Num. of Rules & 10.40 & 13.77 & 39.75 & 131.92 & 16.04 & 7.12 \\
Conditions/Rule & 2.28 & 1.94 & 5.75 & 8.10 & 1.96 & 2.23 \\
Rulebase Size &  23.71 & 26.71 & 228.56 & 1068.55 & 31.43 & 15.87 \\
\bottomrule
\end{tabular}}
\end{table}

FGRT produces models with an average of 10.40 rules and 2.28 conditions per rule, yielding a total rulebase size (rules × conditions/rule) of 23.71. This complexity is comparable to FRR (26.71) and RIPPER (31.43), and substantially smaller than CART (228.56) and especially C4.5 (1068.55). While FGA produces the smallest rule bases (15.87), FGRT's slightly larger models achieve competitive or superior accuracy on most datasets (7 out of 10 datasets where FGRT$^*$ matches or exceeds FGA performance).

\subsection{Hyperparameter study}

\begin{figure*}
    \centering
    \includegraphics[width=\linewidth]{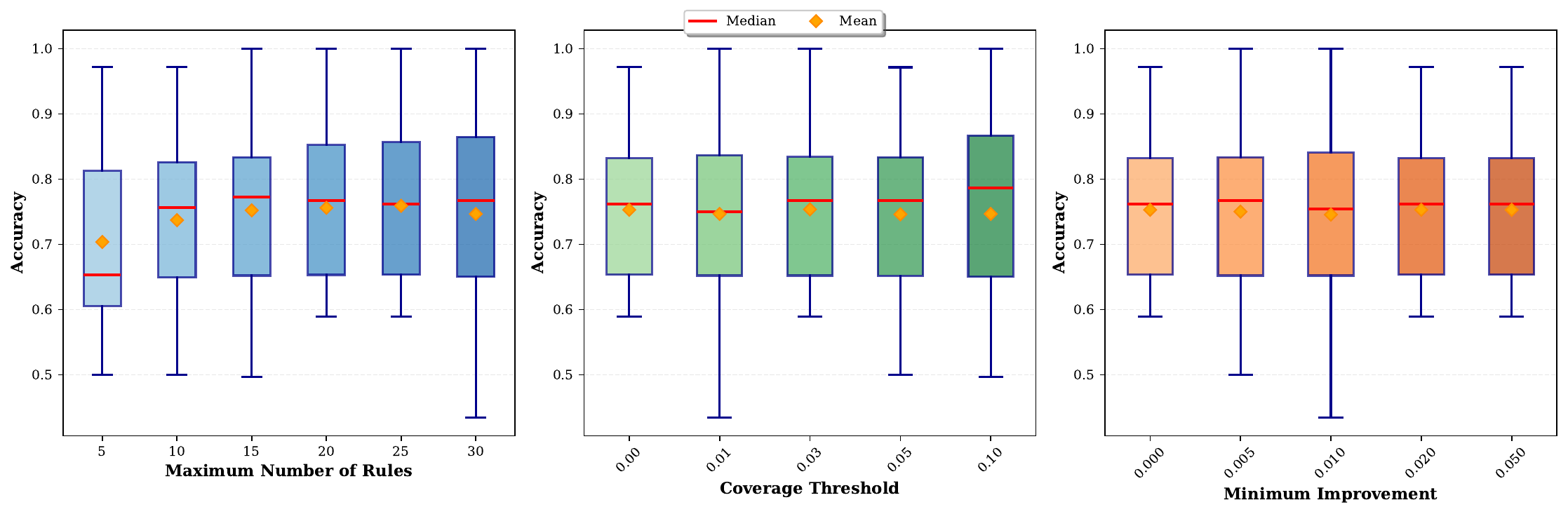}
    \caption{Averaged performance for all datasets using different configuration hyperparameters.}
    \label{fig:ablation}
\end{figure*}

Figure~\ref{fig:ablation} presents an ablation study examining the sensitivity of FGRT to its three main hyperparameters across all datasets. On the left, it illustrates the evolution of FGRT complexity and performance. Performance clearly improves as the number of rules increases from 5 to 15. Beyond this point, however, we observe diminishing returns, with performance even decreasing at 30 rules. This shows that FGRT performs well with a small number of rules, but scaling to larger rulebases requires a more sophisticated search or some refinement of the fuzzy partitions to properly exploit larger numbers of rules. On the center, coverage threshold exhibits minimal impact on accuracy, with performance remaining stable across the tested range (0.00 to 0.10), though a slight improvement is observed at the highest threshold value. Finally, the minimum improvement threshold shows relatively flat performance across values from 0.000 to 0.050, indicating that FGRT is robust to this parameter choice. These results demonstrate that while the maximum number of rules is a critical hyperparameter requiring careful tuning, FGRT maintains stable performance across different coverage and minimum improvement settings.

\subsection{Running times}
\begin{figure}[htbp]
    \centering
    \includegraphics[width=0.53\linewidth]{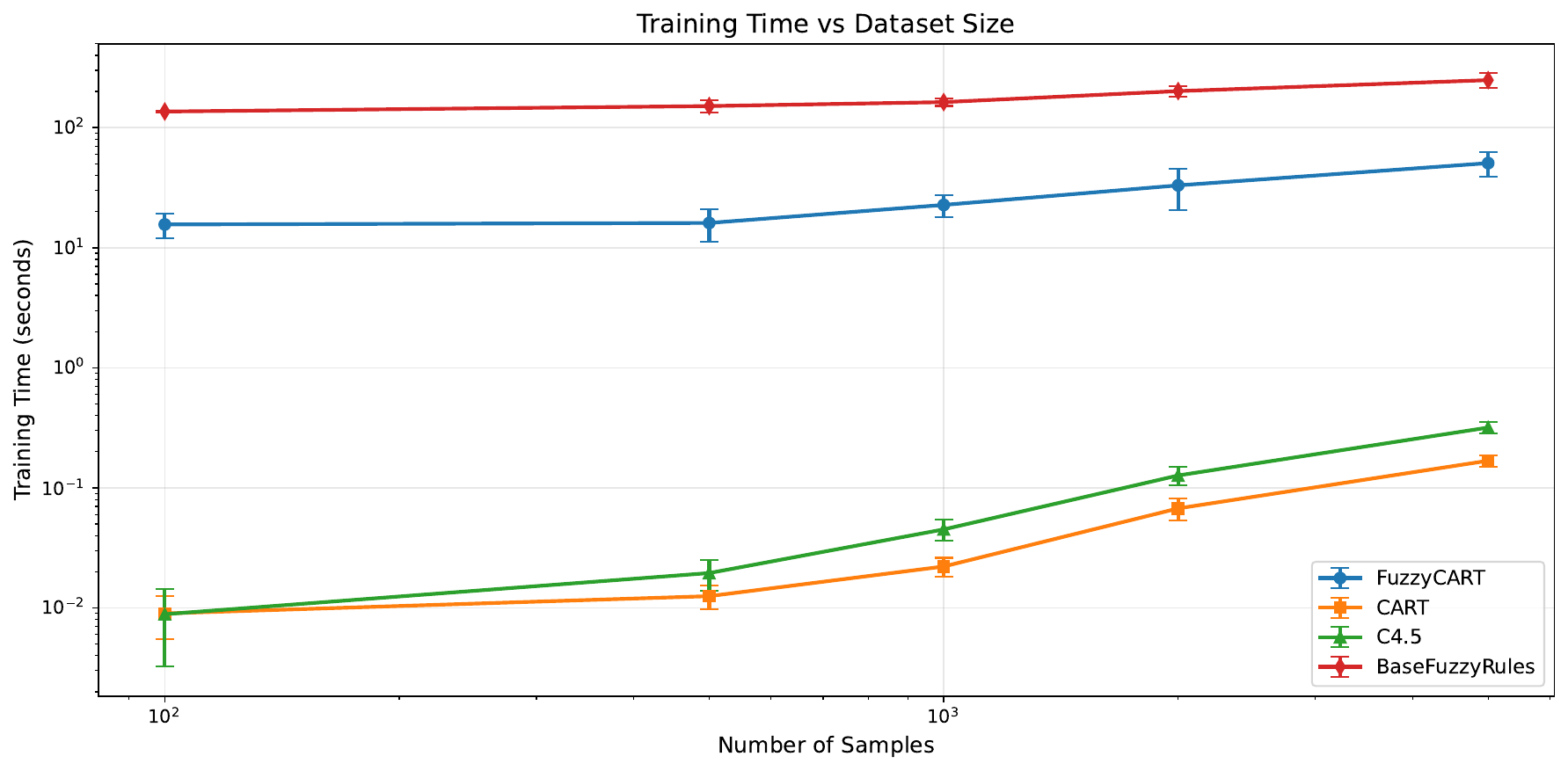}
    \includegraphics[width=0.45\linewidth]{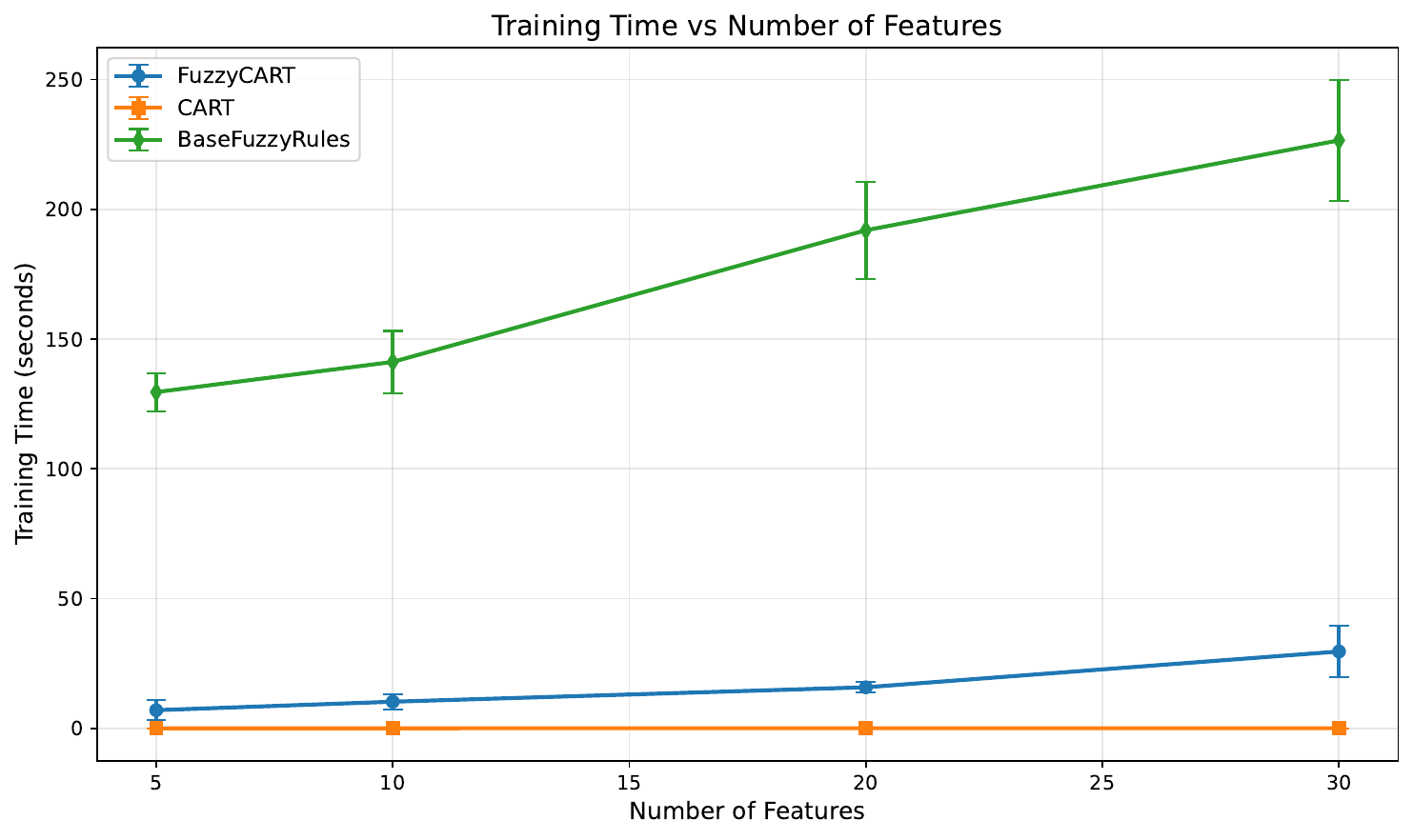}
    \caption{Comparison of running times between the genetic fine-tuning algorithm and our greedy approach, averaged for all the tested datasets. Left shows the evolution for a growing number of samples, and right for a larger number of features}
    \label{fig:runtime_comparison}
\end{figure}

The computational running times of CART, C4.5, the genetic FRBC and our proposed FGRT are presented in Fig. \ref{fig:runtime_comparison}. demonstrates a compelling advantage of the FuzzyCART approach over genetic fine-tuning. In the left panel, as dataset size increases from 10$^2$ to 10$^4$ samples, FuzzyCART maintains training times below 10$^2$ seconds, while BaseFuzzyRules requires over 10$^4$ seconds—representing a speedup of more than two orders of magnitude. Similarly, the right panel shows that FuzzyCART's computational cost grows better with increasing feature dimensionality.

\section{Conclusions}\label{sec:conclusions}
In this work, we propose a greedy method for training fuzzy rule-based classifiers using linguistic partitions and fuzzy Gini impurity reduction. We introduce a lightweight partition optimization framework that tunes membership functions to maximize class separability, using a novel encoding scheme that guarantees validity by construction. Our approach achieves computational efficiency, interpretability preservation, and adaptive partition improvement.

Experiments on 10 benchmark datasets show that our proposed method, FGRT, achieves 76.92\% mean accuracy with prestablished fuzzy partitions, competitive with genetic search and with stablished methods like RIPPER (76.93\%). With optimized partitions, accuracy improves to an average 79.67\%, approaching the FRR (81.11\%) while maintaining low computational complexity. 
FGRT also produces models with only 10.40 rules on average—approximately 4-13× fewer than CART (39.75) and C4.5 (131.92), which makes it a compelling alternative for practicioners.

For future research, we plan to explore joint optimization across non-independent features and alternative optimization objectives to incorporate complexity penalties in the FGRT construction.

\section{Acknowledgement}

This research and Javier Fumanal-Idocin were supported by EU Horizon Europe under the Marie Skłodowska-Curie COFUND grant No 101081327 YUFE4Postdocs. Raquel Fernandez-Peralta is funded by the EU NextGenerationEU through the Recovery and Resilience Plan
for Slovakia under the project No. 09I03-03-V04- 00557.

\bibliographystyle{ieeetr}
\bibliography{aaai2026}

@article{belhadj2021fuzzy,
  title={Fuzzy Version of Gini’s Index},
  author={Belhadj, Besma and Kaabi, Firas and Bouanani, Mejda},
  journal={Social Indicators Research},
  volume={157},
  number={3},
  pages={1079--1087},
  year={2021},
  publisher={Springer}
}

@article{rudin2019stop,
  title={Stop explaining black box machine learning models for high stakes decisions and use interpretable models instead},
  author={Rudin, Cynthia},
  journal={Nature machine intelligence},
  volume={1},
  number={5},
  pages={206--215},
  year={2019},
  publisher={Nature Publishing Group UK London}
}

@book{mendel2024explainable,
  title={Explainable Uncertain Rule-Based Fuzzy Systems},
  author={Mendel, Jerry M},
  year={2024},
  publisher={Springer}
}

@book{breiman2017classification,
  title={Classification and regression trees},
  author={Breiman, Leo and Friedman, Jerome and Olshen, Richard A and Stone, Charles J},
  year={2017},
  publisher={Chapman and Hall/CRC}
}

@book{molnar2020interpretable,
  title={Interpretable machine learning},
  author={Molnar, Christoph},
  year={2020},
  publisher={Lulu. com}
}

@inproceedings{fernandez2025crisp,
  author={Fernandez-Peralta, Raquel and Fumanal-Idocin, Javier and Andreu-Perez, Javier},
  booktitle={2025 IEEE International Conference on Fuzzy Systems (FUZZ)}, 
  title={Crisp Complexity of Fuzzy Classifiers}, 
  year={2025},
  volume={},
  number={},
  pages={1-6}
}

@article{alcala2011fuzzy,
	title={A fuzzy association rule-based classification model for high-dimensional problems with genetic rule selection and lateral tuning},
	author={Alcal{\'a}-Fdez, Jes{\'u}s and Alcala, Rafael and Herrera, Francisco},
	journal={IEEE Transactions on Fuzzy systems},
	volume={19},
	number={5},
	pages={857--872},
	year={2011},
	publisher={IEEE}
}

@article{wang2021scalable,
	title={Scalable rule-based representation learning for interpretable classification},
	author={Wang, Zhuo and Zhang, Wei and Liu, Ning and Wang, Jianyong},
	journal={Advances in Neural Information Processing Systems},
	volume={34},
	pages={30479--30491},
	year={2021}
}

@article{fumanal2024ex,
  title={Ex-Fuzzy: A Library for Symbolic Explainable Ai Through Fuzzy Logic Programming},
  author={Fumanal Idocin, Javier and Andreu-Perez, Javier},
  journal={Neurocomputing},
  year={2024}
}

@article{fumanal2025compact,
  title={Compact Rule-Based Classifier Learning via Gradient Descent},
  author={Fumanal-Idocin, Javier and Fernandez-Peralta, Raquel and Andreu-Perez, Javier},
  journal={arXiv preprint arXiv:2502.01375},
  year={2025}
}

@book{quinlan2014c4,
  title={C4. 5: programs for machine learning},
  author={Quinlan, J Ross},
  year={2014},
  publisher={Elsevier}
}

@article{jang1993anfis,
  title={ANFIS: adaptive-network-based fuzzy inference system},
  author={Jang, J-SR},
  journal={IEEE transactions on systems, man, and cybernetics},
  volume={23},
  number={3},
  pages={665--685},
  year={1993},
  publisher={IEEE}
}

@article{oh2004hybrid,
  title={Hybrid genetic algorithms for feature selection},
  author={Oh, Il-Seok and Lee, Jin-Seon and Moon, Byung-Ro},
  journal={IEEE Transactions on pattern analysis and machine intelligence},
  volume={26},
  number={11},
  pages={1424--1437},
  year={2004},
  publisher={IEEE}
}

@article{erion2022cost,
  title={A cost-aware framework for the development of AI models for healthcare applications},
  author={Erion, Gabriel and Janizek, Joseph D and Hudelson, Carly and Utarnachitt, Richard B and McCoy, Andrew M and Sayre, Michael R and White, Nathan J and Lee, Su-In},
  journal={Nature Biomedical Engineering},
  volume={6},
  number={12},
  pages={1384--1398},
  year={2022},
  publisher={Nature Publishing Group UK London}
}

@inproceedings{mctavish2022fast,
	title={Fast sparse decision tree optimization via reference ensembles},
	author={McTavish, Hayden and Zhong, Chudi and Achermann, Reto and Karimalis, Ilias and Chen, Jacques and Rudin, Cynthia and Seltzer, Margo},
	booktitle={Proceedings of the AAAI conference on artificial intelligence},
	volume={36},
	number={9},
	pages={9604--9613},
	year={2022}
}

@inproceedings{qiao2021learning,
	title={Learning accurate and interpretable decision rule sets from neural networks},
	author={Qiao, Litao and Wang, Weijia and Lin, Bill},
	booktitle={Proceedings of the AAAI Conference on Artificial Intelligence},
	volume={35},
	number={5},
	pages={4303--4311},
	year={2021}
}

@article{rudin2022interpretable,
	title={Interpretable machine learning: Fundamental principles and 10 grand challenges},
	author={Rudin, Cynthia and Chen, Chaofan and Chen, Zhi and Huang, Haiyang and Semenova, Lesia and Zhong, Chudi},
	journal={Statistic Surveys},
	volume={16},
	pages={1--85},
	year={2022},
	publisher={The American Statistical Association, the Bernoulli Society, the Institute~…}
}

@article{wang2017bayesian,
	title={A bayesian framework for learning rule sets for interpretable classification},
	author={Wang, Tong and Rudin, Cynthia and Doshi-Velez, Finale and Liu, Yimin and Klampfl, Erica and MacNeille, Perry},
	journal={Journal of Machine Learning Research},
	volume={18},
	number={70},
	pages={1--37},
	year={2017}
}

@article{arrieta2020explainable,
	title={Explainable Artificial Intelligence (XAI): Concepts, taxonomies, opportunities and challenges toward responsible AI},
	author={Arrieta, Alejandro Barredo and D{\'\i}az-Rodr{\'\i}guez, Natalia and Del Ser, Javier and Bennetot, Adrien and Tabik, Siham and Barbado, Alberto and Garc{\'\i}a, Salvador and Gil-L{\'o}pez, Sergio and Molina, Daniel and Benjamins, Richard and others},
	journal={Information fusion},
	volume={58},
	pages={82--115},
	year={2020},
	publisher={Elsevier}
}

@article{scikit-learn,
  title={Scikit-learn: Machine Learning in {P}ython},
  author={Pedregosa, F. and Varoquaux, G. and Gramfort, A. and Michel, V.
          and Thirion, B. and Grisel, O. and Blondel, M. and Prettenhofer, P.
          and Weiss, R. and Dubourg, V. and Vanderplas, J. and Passos, A. and
          Cournapeau, D. and Brucher, M. and Perrot, M. and Duchesnay, E.},
  journal={Journal of Machine Learning Research},
  volume={12},
  pages={2825--2830},
  year={2011}
}

@article{chua2023tackling,
  title={Tackling prediction uncertainty in machine learning for healthcare},
  author={Chua, Michelle and Kim, Doyun and Choi, Jongmun and Lee, Nahyoung G and Deshpande, Vikram and Schwab, Joseph and Lev, Michael H and Gonzalez, Ramon G and Gee, Michael S and Do, Synho},
  journal={Nature Biomedical Engineering},
  volume={7},
  number={6},
  pages={711--718},
  year={2023},
  publisher={Nature Publishing Group UK London}
}

@article{lipton2018mythos,
  title={The mythos of model interpretability: In machine learning, the concept of interpretability is both important and slippery.},
  author={Lipton, Zachary C},
  journal={Queue},
  volume={16},
  number={3},
  pages={31--57},
  year={2018},
  publisher={ACM New York, NY, USA}
}

@inproceedings{cohen1995fast,
	title={Fast effective rule induction},
	author={Cohen, William W and others},
	booktitle={Proceedings of the twelfth international conference on machine learning},
	pages={115--123},
	year={1995}
}

@article{zadeh1975concept,
  title={The concept of a linguistic variable and its application to approximate reasoning},
  author={Zadeh, Lotfi Asker},
  journal={Information sciences},
  volume={8},
  number={3},
  pages={199--249},
  year={1975},
  publisher={Elsevier}
}

@article{wang2023learning,
	title={Learning interpretable rules for scalable data representation and classification},
	author={Wang, Zhuo and Zhang, Wei and Liu, Ning and Wang, Jianyong},
	journal={IEEE Transactions on Pattern Analysis and Machine Intelligence},
	year={2024},
	publisher={IEEE}
}

@article{fumanal2023artxai,
	title={Artxai: Explainable artificial intelligence curates deep representation learning for artistic images using fuzzy techniques},
	author={Fumanal-Idocin, Javier and Andreu-Perez, Javier and Cord, Oscar and Hagras, Hani and Bustince, Humberto and others},
	journal={IEEE Transactions on Fuzzy Systems},
	year={2023},
	publisher={IEEE}
}

@article{gacto2011interpretability,
	title={Interpretability of linguistic fuzzy rule-based systems: An overview of interpretability measures},
	author={Gacto, Maria Jose and Alcal{\'a}, Rafael and Herrera, Francisco},
	journal={Information Sciences},
	volume={181},
	number={20},
	pages={4340--4360},
	year={2011},
	publisher={Elsevier}
}

@article{huhn2009furia,
	title={FURIA: an algorithm for unordered fuzzy rule induction},
	author={H{\"u}hn, Jens and H{\"u}llermeier, Eyke},
	journal={Data Mining and Knowledge Discovery},
	volume={19},
	pages={293--319},
	year={2009},
	publisher={Springer}
}

@misc{kelly2025uci,
  author = {Markelle Kelly and Rachel Longjohn and Kolby Nottingham},
  title = {The UCI Machine Learning Repository},
  year = {},
  url = {https://archive.ics.uci.edu}
}

@inproceedings{gonzalez2024rule,
  title={A Rule-Based Approach for Interpretable Intensity-Modulated Radiation Therapy Treatment Selection},
  author={Gonzalez-Garcia, Xabier and Fumanal-Idocin, Javier and Do Rio, Joan M Nunez and Bustince, Humberto},
  booktitle={2024 IEEE International Conference on Fuzzy Systems (FUZZ-IEEE)},
  pages={1--8},
  year={2024},
  organization={IEEE}
}
\end{document}